%% file: ijcai23.tex
\documentclass{article}
\usepackage{ijcai23}

\usepackage{times}
\usepackage{soul}
\usepackage{url}
\usepackage[hidelinks]{hyperref}
\usepackage[utf8]{inputenc}
\usepackage[small]{caption}
\usepackage{graphicx}
\usepackage{amsmath}
\usepackage{amsthm}
\usepackage{booktabs}
\usepackage{subfigure}
\usepackage[switch]{lineno}
\newcommand{\cnn}{\textsl{CNN}}
\newcommand{\fl}{\textsl{FL}}
\newcommand{\gnn}{\textsl{GNN}}

\newcommand{\kd}{\textsl{KD}}

\newcommand{\gat}{\textsl{GAT}}

\newcommand{\fgl}{\textsl{FGL}}
\newcommand{\gcn}{\textsl{GCN}}
\newcommand{\cka}{\textsl{CKA}}
\newcommand{\firstmodule}{{Federated Node Semantic Contrast}}
\newcommand{\secondmodule}{{Federated Graph Structure Distillation}}
\newcommand{\firstmoduleabbrv}{\textsl{FNSC}}
\newcommand{\secondmoduleabbrv}{\textsl{FGSD}}
\newcommand{\ours}{{Federated Graph Semantic and Structural Learning}}
\newcommand{\oursabbrv}{\textsl{FGSSL}}
\usepackage{amssymb}
\usepackage{enumitem}

\usepackage{tcolorbox}
\tcbuselibrary{theorems}
\tcbset{highlight math/.append style={left=0mm,right=0mm,top=0mm,bottom=0mm, colframe=white}}

\usepackage{bm}
\usepackage{multirow}
\usepackage[export]{adjustbox}

\usepackage{colortbl}
\definecolor{lightgray}{gray}{.9}
\definecolor{deepgray}{gray}{.8}

\usepackage{threeparttable}
\newcolumntype{I}{!{\vrule width 1pt}}
\makeatletter
\newcommand{\thickhline}{%
    \noalign {\ifnum 0=`}\fi \hrule height 1pt
    \futurelet \reserved@a \@xhline
}
\makeatother
\definecolor{mygray}{gray}{.9}
\usepackage{float}
\usepackage[ruled, vlined]{algorithm2e}
\usepackage{pifont}
\usepackage{ragged2e}
\usepackage{amsfonts}

\usepackage[capitalize]{cleveref}
\crefname{section}{Sec.}{Secs.}
\crefname{table}{Tab.}{Tabs.}
\crefname{proposition}{Prop.}{Props.}

\usepackage{xspace}
\makeatletter
\DeclareRobustCommand\onedot{\futurelet\@let@token\@onedot}
\def\@onedot{\ifx\@let@token.\else.\null\fi\xspace}
\theoremstyle{plain}

\theoremstyle{definition}

\theoremstyle{remark}

\def\eg{\textit{e.g}\onedot}

\makeatother
\title{Federated Graph Semantic and Structural Learning}

\author{
Wenke Huang$^{1\dagger}$
\and
Guancheng Wan$^{1\dagger}$
\and
Mang Ye$^{1,2*}$\And
Bo Du$^{1,2}$
\affiliations
$^{1}$School of Computer Science, Wuhan University, Wuhan, China\\
$^{2}$ Hubei Luojia Laboratory, Wuhan, China\\
\emails
\{wenkehuang, guanchengwan, yemang, dubo\}@whu.edu.cn
}

\begin{document}

\maketitle

\renewcommand{\thefootnote}{\fnsymbol{footnote}} %将脚注符号设置为fnsymbol类型，即特殊符号表示
\footnotetext[1]{Corresponding Author}
\footnotetext[2]{Equal Contribution}

\begin{abstract}

% The superiority of graph neural networks (\gnn{}) in modeling graph data has been widely demonstrated. 
Federated graph learning collaboratively learns a global graph neural network with distributed graphs, where the non-independent and identically distributed property is one of the major challenge. 
Most relative arts focus on traditional distributed tasks like images and voices, incapable of the graph structures. 
This paper firstly reveals that local client distortion is brought by both node-level semantics and graph-level structure.
First, for node-level semantic, we find that contrasting nodes from distinct classes is beneficial to provide a well-performing discrimination. 
We pull the local node towards the global node of the same class and push it away from the global node of different classes. 
Second, we postulate that a well-structural graph neural network possesses similarity for neighbors due to the inherent adjacency relationships. However, aligning each node with adjacent nodes hinders discrimination due to the potential class inconsistency.
We transform the adjacency relationships into the similarity distribution and leverage the global model to distill the relation knowledge into the local model, which preserves the structural information and discriminability of the local model. 
Empirical results on three graph datasets manifest the superiority of the proposed method over counterparts.

\end{abstract}

\input{main.tex}
\newpage
% \clearpage
\section*{Contribution Statement}
Wenke Huang and Guancheng Wan contributed equally to this work.
\bibliographystyle{named}

\bibliography{reference}

% \clearpage
\appendix

\input{appendix.tex}

\end{document}

%% file: main.tex
\definecolor{DarkBlue}{RGB}{64,101,149}
\definecolor{azure}{rgb}{0.0, 0.5, 1.0}
\definecolor{gray}{rgb}{0.3, 0.3, 0.3}
\definecolor{DarkGreen}{RGB}{42,110,63}

\newcommand{\tworowbestreshl}[2]{
\begin{tabular}{@{}c@{}}
\textbf{#1}  \\
\fontsize{6.5pt}{1em}\selectfont\color{DarkBlue}{$\uparrow$ \textbf{#2}}
\end{tabular}
}

\newcommand{\tworowsecondreshl}[2]{
\begin{tabular}{@{}c@{}}
\underline{#1}  \\
\fontsize{6.5pt}{1em}\selectfont\color{DarkBlue}{$\uparrow$ \textbf{#2}}
\end{tabular}
}

\newcommand{\tworowreshl}[2]{\begin{tabular}{@{}c@{}}
{#1}  \\
\fontsize{6.5pt}{1em}\selectfont\color{DarkBlue}{$\uparrow$ \textbf{#2}}
\end{tabular}}

\newcommand{\reshl}[2]{
{#1} \fontsize{6.0pt}{1em}\selectfont\color{DarkGreen}{$\!\pm\!$ {#2}}
}

\newcommand{\ourreshl}[2]{
\textbf{}{#1} \fontsize{6.0pt}{1em}\selectfont\color{DarkGreen}{$\!\pm\!$ {#2}}
}

% 标记十字星标
\newcommand{\itemwithdag}[2]{\begin{tabular}{@{}c@{}}
\fontsize{6.5pt}{1em}{({#2}$^\dag$)} {#1}
\end{tabular}}

\section{Introduction}

Federated learning (\fl{}) has shown considerable potential in collaborative machine learning across distributed devices without disclosing privacy~\cite{FGGP_AAAI24}. Although \fl{} has attracted wide research interest and witnessed remarkable progress~\cite{FLSurveyandBenchmarkforGenRobFair_arXiv23}, most of them focus on the tasks like images and voices on the basis of \cnn{} and transformer \cite{ResNet_CVPR16,transformer_NeurIPS17}.
However, many real-world applications generate structured graphical data (\eg, knowledge graph, social network {\cite{liu2024review}}), consisting of vertices and edges~\cite{pandemic_AAAI21}, while \cnn{} and transformer can not deal with them effectively due to the inability to capture the topological structure  {\cite{GCN_ICLR17}}. For these graph applications, graph neural networks (\gnn{}) have won praise for their impressive performance {\cite{S3GCL_ICML24}} because they utilize both the independence of nodes and the unique structure of graphs to mine graph data. Therefore, for the purpose of handling the graph data across multiple participants with growing privacy concerns, Federated Graph Learning (\fgl{}) has become a promising paradigm {\cite{fgl-survey_arxiv22}}.

\begin{figure}[t]
	\begin{center}
    \includegraphics[width=0.9\linewidth]{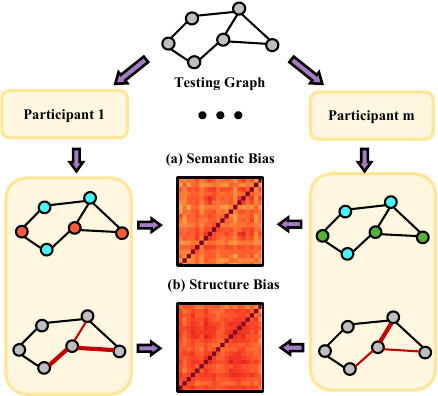}
	\end{center}
	\vspace{-5pt}
	\captionsetup{font=small}
	\caption{\small \textbf{Problem illustration.} We present the structure and semantic level similarities among clients. The deeper color suggests a more similar representation of node and graph across different participants, while the shallower mean dissimilarity.
	(a) Semantic bias: clients show the inconsistency predicted class of node. 
	(b) Structure bias: clients hold distinct similarities among neighborhood nodes.
% 	the different width of edges mean that neighborhoods similarities \textbf{bias} across clients caused by structure heterogeneity. 
% 	node logits difference across clients caused by node feature heterogeneity.  
	In this work, we conduct semantics-level and structure-level calibration to achieve better federated graph learning 
 performance.
 }
	%\caption{\small Visualization of features learned by OCR~\cite{yuan2020object} and Ours on Cityscapes, under the supervision of $\mathcal{L}^{\text{CE}}$ (Eq.~\eqref{eq:CE}) and $\mathcal{L}^{\text{SEG}}$ (Eq.~\eqref{eq:com}), respectively. The points with different colors denote features from different classes. Our approach leads to a well-structured semantic space, showing better intra-class compactness and inter-class separability.}
	\vspace{-5pt}
	\label{fig:problem}
\end{figure}

Notably, data heterogeneity has become an inescapable problem in Federated Learning  {\cite{Advances_arXiv19,FCCL_CVPR22}}. Specifically, the data distribution among different parties presents non-IID (identically and independently distributed) property, which results in the divergence of local direction {\cite{FPL_CVPR23,FSMAFL_ACMMM22}}. Although existing methods have made efforts to restrain the local model with respect to the global model, they mainly design for typical data heterogeneity without special consideration for graph structural bias. However, previous work has demonstrated the importance of exploiting structural knowledge {\cite{GOOD-WSDM23,SUBLIME_WWW22}}. In this paper, we preliminarily investigate the unique characteristics of graph heterogeneity and find that there exists node semantic and graph structural bias in \fgl{} setting. In detail, we leverage the Centered Kernel Alignment (\cka{}) {\cite{CKA_ICML19}} to measure the similarity between node representations given by different client model pairs. Especially, for structure heterogeneity, we utilize the Anonymous Walk Embedding (\textsl{AWE}) to generate a representation for the graph and exploit the Jensen-Shannon distance between pair of graphs to measure the discrepancy. We reveal that there exists severe divergence on both node-level semantic and graph-level structure among clients (\cref{fig:problem}). Therefore, a crucial problem for \fgl{} is that \textit{How to calibrate both node-level semantic and graph-level structure bias in Federated Graph Learning?}

For node-level semantic calibration, we argue that a well-discriminative graph neural network contrasts nodes from different classes to provide a clear decisional boundary. Inspired by the success of supervised contrastive learning \cite{SupCon_NeurIPS20,GREET_AAAI23}, we naturally expect to conduct pull-push operations among different classes on local model to acquire a well-performing decisional ability. However, under federated learning, the local \gnn{} model purely optimizes on private data and drifts towards a distinct local minimum, which means solely relying on the guidance signals provided by the local model is confusing and unreliable.  Prior studies \cite{fedgcn_MDPI22}, attempt to reweight local model during the FL aggregation process, but this sheds little light on identity node semantics bias caused by data heterogeneity. In this work, we investigate the node semantic knowledge and calibrate it during the local training process. We propose \firstmodule{} (\firstmoduleabbrv{}), which encourages the query local node representation to be close to \textbf{global} node embeddings within same class and pushes it away from \textbf{global} node embeddings of different classes.

Besides, for graph-level structural calibration, local clients normally possess a graph that is incomplete and is biased to depict the graph structure. Existing works normally focus on reconstructing the graph structure to handle the local structural bias. For example, FedStar \cite{FedStar_AAAI23} decouples the structure information and encodes it in a personalized way, which brings extra model parameters for local updating. In this work, without more communication cost, we take a free ride to convert stiff graph structure reconstruction into structural relationship maintenance via the given global model during local training. We introduce \secondmodule{} (\secondmoduleabbrv{}). In detail, for each node, we leverage the global model to calculate the similarity of each node with its neighborhoods, based on the adjacency matrix. Then, we require the local model to generate the adjacent node similarity and mimic the global one, which leverages the global model to provide beneficial structural knowledge.

In a nutshell, we propose a novel \ours{} method (\oursabbrv{}). Our contributions are summarized as follows:
% Experimental results on multiple benchmark datasets show the superiority of \secondmoduleabbrv{} in terms of both performance and efficiency. 
% In a nutshell, our contributions are summarized below.
\begin{itemize}[leftmargin=*]
	\setlength{\itemsep}{0pt}
	\setlength{\parsep}{-2pt}
	\setlength{\parskip}{-0pt}
	\setlength{\leftmargin}{-15pt}
	\vspace{-4pt}
    \item We are the first in \fgl{} to decouple the data heterogeneity setting to node semantic level and graph structural level bias respectively. From this perspective, we can ameliorate final degraded performance by calibrating the local training drift, which sheds good light on future research in solving the non-IID problem in \fgl{} scenarios.
    
    % \item We develop methods for both node and graph-level calibration, termed as \firstmodule{} and \secondmodule{} respectively. The former calibrates local node semantics bias by pulling it to the instances given by the global model, while the latter novelly transform the adjacency relationships into the similarity distribution from global model and distill the them into local model. Combining them, we propose a novel federated graph learning frame, namely \oursabbrv{}, which can calibrate bias on both levels and obtain a better final model. 

    \item We introduce a novel federated graph learning (\oursabbrv{}) frame for both node and graph-level calibration. The former \firstmodule{} calibrates local node semantics with the assistance of the global model without compromising privacy. The latter \secondmodule{} transforms the adjacency relationships from the global model to the local model, fully reinforcing the graph representation with aggregated relation.
    
    % while the latter novelly transform the adjacency relationships into the similarity distribution from global model and distill the them into local model. Combining them, we propose a novel federated graph learning frame, namely \oursabbrv{}, which can calibrate bias on both levels and obtain a better final model. 
% \oursabbrv{} can calibrate bias on both levels and enables the final model to learn more informative representations from other clients to achieve better performance.
    
    \item We conduct extensive experiments on benchmark datasets to verify that \oursabbrv{} achieves superior performance over related methods. Taking a free ride with the global model, it does not introduce additional communication rounds and shows stronger privacy since it does not require additional shared sensitive prior information. 
\vspace*{-2pt}
\end{itemize}
 
\section{Related Work}
% Our research is based on previously published works in federated graph learning, contrastive learning on graphs, and knowledge distillation. The most pertinent works are covered as follows.
\subsection{Federated Graph Learning}
Federated graph learning (\fgl{}) facilitates the distributed training of graph neural networks (\gnn{}). Previous literature on \fgl{} can be categorized into two types: inter-graph and intra-graph. Inter-graph \fgl{} involves each participant possessing a set of graphs and collectively participating in federated learning (\fl{}) to improve the modeling of local data or generate a generalizable model \cite{fedGCL_NeurIPS21}.  In contrast, intra-graph \fgl{} involves each participant owning only a subset of the entire graph and the objective is to address missing links \cite{fedsage+} or discover communities \cite{community_arxiv22}. 
However, both of them are confronted with the non-IID issue which degrades the collaboratively learned model performance. Conventional methods solving the non-IID in \fl{} field (\eg, FedProx \cite{fedprox} and MOON \cite{MOON_CVPR21}) meet the absence of design for \fgl{} scenarios. Some preceding methods are dedicated to handling the non-IID problem for \fgl{}. FedGCN {\cite{fedgcn_MDPI22}} tries to reweight local model parameters via an attention mechanism. FILT+ {\cite{filt+_arxiv21}} pulls the local model closer to the global model by minimizing the loss discrepancy between a local model and the global model. However, they focus on leveraging the issue from model respect and fail to effectively exploit the unique characteristics of the graph data. In this paper, we consider inter-graph \fgl{} and deal with the non-IID via exploiting the graphic characteristics and decoupling into node-level semantic and graph-level structure calibration.

\subsection{Contrastive Learning on Graphs}
In recent years, contrastive learning has seen a resurgence of interest in the field of visual representation learning {\cite{moco_CVPR20,SimCLR_ICML20}}. This success has spurred a wealth of research exploring the adaptation of contrastive learning to graph-like data for self-supervised methods {\cite{gcl-survey_arxiv21,gcl-survey_IEEE22}}. Traditional unsupervised methods on graph representation learning approaches {\cite{node2vec_kdd16,deepwalk_kdd14}}, adhere to a contrastive structure derived from the skip-gram model. The graph autoencoder (GAE) {\cite{GAE_arxiv16}} is a self-supervised learning technique that aims to reconstruct the graph structure while The MVGRL {\cite{MVGRL_ICML20}} intends to do node diffusion and compare node representation to augmented graph representation in order to learn both node-level and graph-level representation. Similar to SimCLR {\cite{SimCLR_ICML20}}, GRACE {\cite{GRACE_ICML20}} constructs two augmented views of a graph by randomly perturbing nodes and edges, and subsequently learns node representations by pushing apart representations of every other node while bringing together representations of the same node in the two different augmented graphs within the same network. Apart from self-supervised tasks, SupCon {\cite{SupCon_NeurIPS20}} firstly extend the self-supervised batch contrastive approach to the fully-supervised setting. In this work, we examine the contrastive method in distributed systems and conduct a inter-view based contrast between the global and local models respectively. Moreover, we consider the supervised contrast that leveraging the label as a signal to choose positive samples for calibrating the node embedding to be more similar to the global node embedding.

\subsection{Knowledge Distillation}
Knowledge Distillation (\kd{}) {\cite{KD_arXiv15}} is a technique that has been extensively studied and applied in various areas of machine learning, including image classification, natural language processing, and graph representation learning. The key aspect of \kd{} is transferring knowledge from a complex and powerful teacher model to a more limited student model. In many works, knowledge distillation is typically used to train a smaller student network under the guidance of a larger teacher network with minimal to no performance degradation {\cite{KD_PAMI_21}}. In practice, knowledge distillation forces the feature or logit output of the student network to be similar to that of the teacher network. Researchers have attempted to improve knowledge distillation methods by introducing new techniques such as model distillation \cite{modeldiss_ECCV19}, feature distillation \cite{Fitnets_ICLR15}, and relation distillation \cite{RKD_CVPR19}. In this work, we focus on ameliorating the heterogeneity of graph structure by adapting relation-based \kd{} techniques for the \fgl{} domain. We first transform the adjacency relationships into similarity distribution from global view, then distill them into the local model. In this way, we leverage aggregated contextual neighborhood information from global view and calibrate the drift caused by graph structure from the locally biased data.

\section{Methodology}
% In this section, we introduce preliminaries of federated graph learning in \cref{sec:preliminary} and further illustrate the motivation of our work in \cref{sec:motivation}. Then, we introduce proposed solution in \cref{sec:method}.

\subsection{Preliminaries}
\label{sec:preliminary}
\textbf{Graph Neural Newrok}.
Graph neural networks (\gnn{}), \eg, graph convolutional networks (\gcn{}) {\cite{GCN_ICLR17}} and Graph Attention Networks (\gat{}) (\cite{GAT_arxiv17}), improved the state-of-the-art in informative graph data with their elegant yet powerful designs. In general, given the structure and feature information of a graph $\mathcal{G} = (V, A, X)$, where $V$, $A$, $X$ denote nodes, adjacency matrix and node feature respectively, \gnn{} targets to learn the representations of graphs, such as 
the node embedding $h_i \in \mathbb{R}^{d}$. A \gnn{} typically involves two steps: the processes of message propagation and neighborhood aggregation. In this process, each node in the graph iteratively collects information from its neighbors with its own information in order to update and refine its representation. Generally, an $L$-layer \gnn{} can be formulated as 
\begin{equation}
    h^{(l+1)}_i = \sigma (h^{(l)}_i, \mathit{AGG}(\{h^{(l)}_j; j \in A_i\})), \forall l\in [L],
\end{equation}
where $h^{(l)}_{i}$ denotes the representation of node $v$ at the $l^{th}$ layer, and $h^{(0)}_{i}= v_i$ represents the node feature. $A_i$ is defined as the neighbors of node $v_i$, $\mathit{AGG(\cdot)}$ is a aggregation function that can vary for different \gnn{} variants, and $\sigma$ means a activation function.

After L message-passing layers, the final node embedding $h_i$ is passed to a project head $F$ to obtain logits:
\begin{equation}
    z_i = F(h_i).
\end{equation}
In this paper, we examine proposed \oursabbrv{} in node-level tasks (\eg, node classification), and $F$ is defined as the classifier head. Specially, we utilize L-1 layers as \gnn{} feature extractor and the L layer as $F$.

\subsubsection{Centralized Aggregation}
In vanilla FL setting there is always a central server with $M$ clients, the $m\text{-th}$ client owns a private dataset $D^m$ and $|D|$ is the total size of samples over all clients. FedAvg {\cite{FedAvg_AISTATS17}} is a foundational algorithm in the field of federated learning, which serves as a starting point for the design of more advanced \fl{} frameworks. It operates by aggregating the updated model parameters from individual clients and redistributing average of these parameters back to all clients:
% \begin{equation}
% \label{eq:fedavg}
%     \theta^{(r+1)} = \sum _{m=1}^{M} \frac{|D_i|}{|D|} w_m^{(r)}.
% \end{equation}
\begin{equation}
\label{eq:fedavg}
    {\theta} \leftarrow \sum_{m=1}^{M} \frac{\left|D^{m}\right|}{|D|} \theta^{m}.
\end{equation}
% Specifically, at each communication round $t$, clients $m$ train the model provided by the server using its own local data $\mathcal{D}_i$ for $R$ epochs. The client then sends its updated model parameters back to the server, which aggregates and averages the received parameters before redistributing them to all client devices. This process is repeated until the model has converged or a predetermined number of communication rounds has been reached. The client $m$ then transmits its updated parameters $\theta^{(r)}_{i}$ to the server, and the server will aggregate these updates by  \cref{eq:fedavg}. Then, the server broadcasts the new parameters $\theta^{(r+1)}$ to remote clients, and at the $(r+1)$ round clients use $\theta^{(r+1)}$ to start for another $E$ epochs of their local training.
In this study, we utilize the Federated Learning (\fl{}) framework to enable collaborative learning on isolated graphs among multiple data owners, without the need to share raw graph data. By doing so, we aim to obtain a global node classifier. Specifically, when model parameters are set to $\theta$ for the Graph Neural Network (\gnn{}) encoder and classifier $F$, we formalize the global objective:
\begin{equation}
\label{eq:go}
    \arg \min\frac{1}{M} \sum_{m}^{M} \mathcal{L}^{m}(\theta^m;D^m).
\end{equation}
Normally, the loss function $\mathcal{L}^m$ in \cref{eq:ce} is cross-entropy loss as each node which is optimized with softmax operation:
\begin{equation}\small\label{eq:ce}
% \mathcal{L}^{\textsl{CE}}=- \frac{1}{|D^m|} \sum_{i=1}^{|D^m|}\bm{1}_{c}\log (\texttt{softmax}({\bm{z_i}})),
 \mathcal{L}_i^{\textsl{CE}}=-\bm{1}_{c_i}\log (\texttt{softmax}({\bm{z_i}})),
\vspace{-3pt}
\end{equation}
where $\bm{1}_{c_i}$ denotes the one-hot encoding of the label $c_i$. 
% Given ${\bm{z}}$ for node $i$ with its ground truth label $c\!\in\!\mathcal{C}$, the logarithm is defined as element-wise, and $\texttt{softmax}({z}_c)\!=\!\frac{\exp({z}_c)}{\sum_{c'\!=\!1}^{|\mathcal{C}|}\exp(z_{c'})}$. 

\subsection{Motivation}
\label{sec:motivation}
Commonly,  federated graph learning aims at training a shared global \gnn{} model, where clients have their own graphs and do not expose private data. In real-world applications,  heterogeneous data distribution exists among clients. Therefore, clients present divergent optimization directions, which impair the performance of the global \gnn{} model. We also show that this client divergence manifests in node-level semantics and graph-level structure aspects.
We leverage the pairwise Centered Kernel Alignment (\cka{}) {\cite{CKA_ICML19}} and calculate the similarity between arbitrary \gnn{} models on the same input testing samples. \cka{} generates the similarity score ranging from 0 (not at all similar) to 1 (identical). We select 20 clients and train the local \gnn{} model for 100 epochs, simultaneously taking the node output from different models as node representation. As shown in \cref{fig:problem}, considering both node semantics and graph structure calibration into account is beneficial to learning a better shared \gnn{} model.

\subsection{Proposed Method}
\label{sec:method}

\textbf{\firstmodule{}}. 
% \subsubsection{Discriminativeness of Feature Space}
Generally, the goal of node classification is to identify all samples. Thus, the \gnn{} module should maintain the discernible patterns. Inspired by the success of supervised contrastive learning, we naturally expect to contrast the node features of different classes. For the local model, we pull the node feature vectors closer to the positive samples from the same semantics and push them far away from negativeness with distinct classes. 
Specifically, for the node $v_i$, its embedding $h^m_i$ generated by local \gnn{} encoder $G^m(\cdot)$ with its ground truth $c_i$, the positive samples are other nodes  belonging to the same class $c_i$, while the negatives are the nodes from the different classes $\mathcal{C}\backslash c_i$. Our supervised, local node-wise contrastive loss is defined as:
\begin{equation} 
\small
 \mathcal{L}_{i}^{CON}=\frac{-1}{\left|\boldsymbol{P}_{i}\right|} \sum_{p \in \boldsymbol{P}_{i}}\log \frac{{\varphi({h}^m_{i}, {h}^m_{p},\tau) }} {{\varphi({h}^m_{i}, {h}^m_{p},\tau)} + \sum_{{k}\in \boldsymbol{K}_{i}} {\varphi({h}^m_{i}, {h}^m_{k},\tau) }}, 
\end{equation}
where $\boldsymbol{P}_i$ and $\boldsymbol{K}_i$ denote the collections of the positive and negative samples sets for the node $v_i$. We define the $\tau$ as a contrastive hyper-parameter and $\varphi$ is formulated as: 
\begin{equation}
\begin{split}
\varphi(h_{i},h_{j},\tau) = 
exp( \frac{h_{i} \cdot h_{j}}{||h_{i}|| \ ||h_{j}||}/\tau).
\end{split}
\end{equation}
However, it is widely known that private models present drift from the ideal global optima. Thus, naively leveraging the private model to provide the positive and negative sets would further skew the local optimization direction. In our work, we argue that the shared global model aggregates knowledge from multiple parties and presents less bias than the local model. In this paper, we propose \firstmodule{} (\firstmoduleabbrv{}), which leverages the global model to provide positive and negative cluster representations for each local node embedding. We further reformulate the aforementioned supervised node contrastive learning as follows:
\begin{equation}\small 
\label{eq:fnsc}
 \mathcal{L}_{i}^{\firstmoduleabbrv{}}=\frac{-1}{\left|\boldsymbol{P}_{i}\right|} \sum_{p \in \boldsymbol{P}_{i}}\log \frac{{\varphi({h}^m_{i}, {h}^g_{p},\tau) }} {{\varphi({h}^m_{i}, {h}^g_{p},\tau)} + \sum_{{k}\in \boldsymbol{K}_{i}} {\varphi({h}^m_{i}, {h}^g_{k},\tau) }},
\end{equation}
% For each communication round $\methcal{R}$, we duplicate the global model parameters and preserve a global model for local training. In this section, 
where $h^g$ denotes the the node embedding generated by the \gnn{} encoder $G^g(\cdot)$
% We leverage the shared \gnn{} encoder $G^g(\cdot)$ to generate the node embeddings from the global view.
Moreover, given the node embedding $h_i^m$ generated by local \gnn{} encoder $G^m(\cdot)$, we pull the node $v_i$ from local view and its pairwise one $h_i^g$ in global view together, simultaneously pull it and nodes from global view with the same class $c_i$ together.  

Notably, the recent success of contrastive learning in image or video processing is largely due to carefully designed image augmentations \cite{mangye_agu_PAMI20,mangye_cl_CVPR19}. These augmentations allow the model to explore a wider range of underlying semantic information and obtain better performance. In this section, we adopt a similar strategy for graph data by using an augmentation module, denoted by $Aug(\cdot)$, to generate two different views of the graph. Prior research has produced various methods for graph augmentation, which can be divided into two categories: topology (structure) transformation and feature transformation (\eg, Edge Removing and Feature Masking) \cite{gcl-survey_arxiv21,GRACE_ICML20}. In order to enforce local clients to acquire a well-discriminative ability, we leverage both augmentations in our augmentation modules.
Furthermore, we propose an \textbf{asymmetric} design for the contrast process, which utilizes stronger augmentations for the local \gnn{} and weaker augmentations for the global \gnn{}, given by $\widetilde{\mathcal{G}}_1 = Aug_s(\mathcal{G})$ for strong $Aug(\cdot)$ and $\widetilde{\mathcal{G}}_2 = Aug_w(\mathcal{G})$ for weak $Aug(\cdot)$. This would give local clients great strength to optimize towards the global direction, meanwhile, the global model can provide stable contextual semantic information to local training process. We further demonstrate the effectiveness of this asymmetric augmentation strategy in \cref{table:augmentation}.

\begin{figure*}[t!]
	\centering
    \begin{center}
		\includegraphics[width=\linewidth]{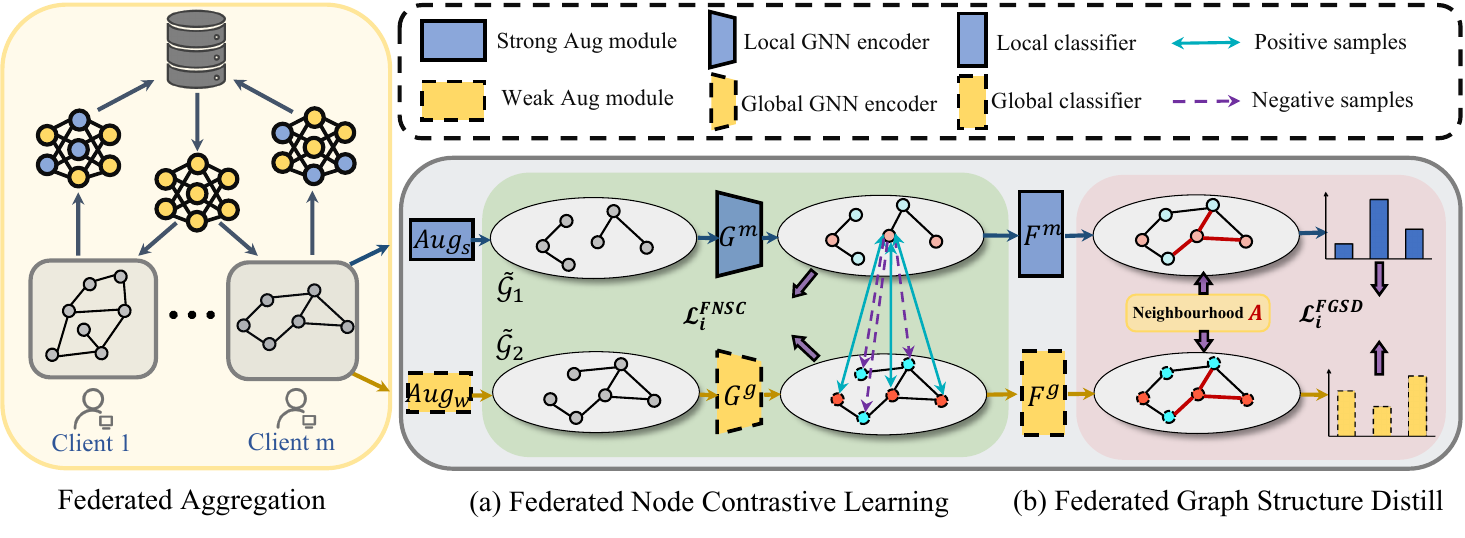}
	\put(-240,75){\scriptsize{\cref{eq:fnsc}}}
    \put(-30,75){\scriptsize{\cref{eq:msd}}}
    \end{center}
    \vspace{-15pt}
    \caption{\textbf{Architecture illustration} of \ours{} (\oursabbrv{}). The left yellow box corresponds to the federated aggregation scheme (\eg FedAvg), while the right grey box suggests the local training process. \oursabbrv{} includes two components: (a) \firstmodule{}   and (b) \secondmodule{}. Best viewed in color. Zoom in for details.
    }
    \label{fig:framework}
    \vspace{-5pt}
\end{figure*}

\noindent \textbf{\secondmodule{}}.
For graph-level calibration, it is normally assumed that adjacent nodes will share similar representations. However, under federated learning, each client fails to effectively depict this relationship because local data is normally incomplete. The straightforward solution is to directly align the query local node feature with the neighborhood nodes from the global model. However, it could potentially disrupt the discriminability because neighboring nodes probably belong to different classes. Motivated by similarity knowledge distillation {\cite{SEED_ICLR21,ISD_ICCV21}}, we propose \secondmodule{} (\secondmoduleabbrv{}) to overcome semantic inconsistency of adjacent nodes, which maintains graphic structure knowledge via the support of the global model. We measure the similarity of the query node with neighboring nodes from the global model output and then optimize the local network to mimic the similarity distribution from the global view. 
Specifically, for the node $v_i$, $z_i$ is the logit output given by the node classifier $F$, we denote the $A_i$ as the neighborhood node set and define the $S^{g}(v_i,A_i)$ as the similarity of the selected node vector with adjacent nodes computed by the \textbf{global} model:
\begin{equation}\small
\begin{split}
S^{g}(v_i,A_i) = [S^{g}_1,\ldots,S^{g}_{|A_i|}],\\
S^{g}_j =\frac{\exp\big(\big(z^{g}_i \cdot {z^{g}_j}^T\big)/\omega \big)}
{\sum_{j \in A_i} \text{exp}\big(\big(z^{g}_i \cdot {z^{g}_j}^T)\big)/\omega \big)},
\label{eq:simdist_global}
\end{split}
\end{equation}
where $\omega$ is the distillation hyper-parameter, $(\ )^T$ means transpose operation. Then, we measure similarity distribution from the $m$ local model,  $S^{m}(v_i,A_i)$, which is formed by:
\begin{equation}\small
\begin{split}
S^{m}(v_i,A_i) = [S^{m}_1,\ldots,S^{m}_{|A_i|}],\\
S^{m}_j =\frac{\exp\big(\big(z^{m}_i \cdot {z^{m}_j}^T\big)/\omega \big)}
{\sum_{j \in A_i} \text{exp}\big(\big(z^{m}_i \cdot {z^{m}_j}^T)\big)/\omega \big)}.
\label{eq:simdist_local}
\end{split}
\end{equation}
The \secondmoduleabbrv{} (\secondmodule{}) loss is calculated as the following:
\begin{equation}\small
\mathcal{L}_i^{\text{\secondmoduleabbrv{}}} =  S^{g}(v_i,A_i) \log \frac{S^{g}(v_i,A_i)}{S^{m}(v_i,A_i)}.
\label{eq:msd}
\end{equation}
Finally, the overall objective to be maximized is then formalized as the average across all nodes of the accumulation of the losses discussed above and is defined by:
\begin{equation}\small
\label{eq:fgsd}
    \mathcal{L}=\frac{1}{N} \sum_{i=1}^{N}\left(\mathcal{L}^{\textsl{CE}}_i + \lambda_C \mathcal{L}_i^{\text{\firstmoduleabbrv{}}} + \lambda_D \mathcal{L}_i^{\text{\secondmoduleabbrv{}}} \right).
\end{equation}
To sum up, \ours{} \oursabbrv{} leverage the global model to simultaneously calibrate local model from the node-level semantics and graph-level structure, which effectively handles the heterogeneous graph data and learns a well-performing global \gnn{} model. We further illustrate the \oursabbrv{} in the \cref{alg:ours}.
% . At each training epoch $E$, we first generate two graph views $\widetilde{\mathcal{G}}_{1}$ and $\widetilde{\mathcal{G}}_{2}$ of graph $G$. Then, we transfer them into local and global model respectively,

% \setlength{\textfloatsep}{0.3cm}
% \setlength{\floatsep}{0.3cm}

\begin{algorithm}[t]
\caption{The \oursabbrv{} Framework}
\label{alg:ours}
\SetAlgoLined
\SetNoFillComment
\SetArgSty{textnormal}
\small{\KwIn{communication rounds $T$, local epochs $E$, participant scale $M$, $m^{th}$ client private graph data $\mathcal{G}^m(V,A,X;Y)$, private model $\theta^m$, temperature $\tau$, distillation parameter $\omega$, loss weight $\lambda_C$ and $\lambda_D$, learning rate $\eta$}}
\small{\KwOut{The final global model $\theta_t$}}

\BlankLine
\For {$t=1, 2, ..., T$}{

    \emph{Participant Side}\;
    
    \For {$m=1, 2, ..., M$ \textbf{in parallel}}{
    send the global model $\theta_t$ to $m\text{-th}$ client
    
    $\theta^m_t \leftarrow$ \textbf{LocalUpdating}($\theta_t$, $m$)
    }
    
    % \BlankLine
    
    \emph{Server Side}\;

    % \BlankLine

    $ \theta_{t+1} \leftarrow \sum_{m=1}^{M} \frac{\left|D^{m}\right|}{|D|} \theta^{m}_t$
}
return $\theta_t$
\BlankLine

\textbf{LocalUpdating}($\theta_t$, $m$):
\BlankLine

\textbf{Initialize} $G^g(\cdot),F^g(\cdot) \leftarrow$ $\theta_t$

\BlankLine
\textbf{Initialize} $G^m(\cdot),F^m(\cdot) \leftarrow$ $\theta_t$
\BlankLine
\textbf{Freeze} $G^g(\cdot),F^g $
\BlankLine
\For{ $e=1, 2, ..., E$}{
    %local global
    $Z = \theta^m(X)$ 

    % \BlankLine

    $ L^{CE} \leftarrow CE(Z,Y)$ in \cref{eq:ce}

    % \BlankLine
    
    % {\color{DarkBlue}{\tcc{Generate Augmented Graph}}}
    $\widetilde{\mathcal{G}}_1, \widetilde{\mathcal{G}}_2 \leftarrow Aug_s(\mathcal{G}), Aug_w(\mathcal{G})$

    % \BlankLine
    
    % {\color{DarkBlue}{\tcc{\firstmodule{}}}}
    $H^m, H^g \leftarrow G^m(\widetilde{\mathcal{G}}_1), G^g(\widetilde{\mathcal{G}}_2)$

    % \BlankLine

    $L^{FNSC} \leftarrow (H^m, H^g)$ through \cref{eq:fnsc} 

    % \BlankLine

    % {\color{DarkBlue}{\tcc{\secondmodule{}}}}
    $Z^m, Z^g \leftarrow F^m(H^m), F^g(H^g)$

    % \BlankLine

    $S^{g}(V,A) \leftarrow (Z^g)$ by \cref{eq:simdist_global}

    % \BlankLine

    $S^{m}(V,A) \leftarrow (Z^m)$ by \cref{eq:simdist_local}

    % \BlankLine

    $L^{FGSD} \leftarrow (S^{m}(V,A), S^{g}(V,A))$ through \cref{eq:fgsd} 
    
    % \BlankLine
    
    % {\color{DarkBlue}{\tcc{Overall Objective}}}
    $ \mathcal{L}=\mathcal{L}^{\textsl{CE}} + \lambda_C \mathcal{L}^{\text{\firstmoduleabbrv{}}} + \lambda_D \mathcal{L}^{\text{\secondmoduleabbrv{}}} $

    % \BlankLine

    $\theta^m \leftarrow \theta^m-\eta \nabla \mathcal{L}$
}
return $\theta^m$

% \BlankLine
% {\textbf{LocalUpdating}($f$)}:

% $f^k \leftarrow f$ 
% % {\footnotesize{\color{azure}{\tcp*{Distribute global model}}}}

% \For {$t=1, 2, ..., T$}{
%     \For{$(x_i, y_i) \in D^k$}{
    
%     $z_i^k = f^k(x_i)$
    
%     $z_i^k = \|z_i^k\| \!\cdot\! \hat{z}_{i}^{k}$
    
%     \BlankLine

%     {\color{DarkBlue}{\tcc{\secondmodule{}}}}
%     $\mathcal{L}_{\secondmoduleabbv{}} \leftarrow (\hat{z}_{i}^{k},\tau) $ in \cref{eq:federatednormce}

%     \BlankLine
%     $f^k  \leftarrow f^k  - \eta \nabla \mathcal{L}_{\secondmoduleabbv{}}$
%     }
% }

% {\color{DarkBlue}{\tcc{Local logits magnitude}}}
% ${\|z^k\|} =\sum_{i=1}^{n^k} \|f^k(x_i)\| / n^k$

% return  $f^k, {\|z^k\|}$

\end{algorithm}

%Experiments

\section{Experiments}
\label{sec:experiments}
\subsection{Experimental Setup}
In this paper, we perform experiments on node-level tasks defined on graph data: we choose node classification to confirm the efficacy of \oursabbrv{} in various testing environments. 
The code is available at {\url{https://github.com/GuanchengWan/FGSSL}}.

\begin{table*}[t]\small
\centering
\scriptsize{
\resizebox{\linewidth}{!}{
		\setlength\tabcolsep{0.5pt}
		\renewcommand\arraystretch{1.1}
\begin{tabular}{r||cccIcccIccc}
\hline\thickhline
\rowcolor{lightgray} & 
\multicolumn{3}{cI}{Cora} & 
\multicolumn{3}{cI}{Citeseer} & 
\multicolumn{3}{c}{Pubmed}
\\
\cline{2-10}
\rowcolor{lightgray}
\multirow{-2}{*}{Methods} 
& $M \! = \! 5$ & $M \! = \! 7$ & $M \! = \! 10$ 
& $M \! = \! 5$ & $M \! = \! 7$ & $M \! = \! 10$ 
& $M \! = \! 5$ & $M \! = \! 7$ & $M \! = \! 10$ 
\\

\hline\hline

Global 
& \multicolumn{3}{cI}{\reshl{87.78}{1.34}} 	
& \multicolumn{3}{cI}{\reshl{76.91}{1.02}} 	
& \multicolumn{3}{c}{\reshl{88.38}{0.33}} 	
\\

Local
& \reshl{61.54}{0.83} & \reshl{45.32}{1.52} & \reshl{32.42}{2.81}	
& \reshl{73.85}{1.20} & \reshl{62.87}{2.45} & \reshl{48.91}{2.34}	
& \reshl{83.81}{0.69} & \reshl{72.34}{0.79} & \reshl{59.19}{1.31}	
\\
\hline

FedAvg
& \reshl{86.63}{0.35} & \reshl{86.21}{0.21} & \reshl{86.01}{0.17}	
& \reshl{76.37}{0.43} & \reshl{76.57}{0.46} & \reshl{75.92}{0.21}
& \reshl{85.29}{0.83} & \reshl{84.27}{0.29} & \reshl{84.57}{0.29}
\\

FedProx
& \reshl{86.60}{0.59} & \reshl{86.27}{0.12} & \reshl{86.22}{0.25}
& \reshl{77.15}{0.45} & \reshl{77.28}{0.78} & \reshl{76.87}{0.80}
& \reshl{85.21}{0.24} & \reshl{84.01}{0.59} & \reshl{84.98}{0.65}	
\\

FedOpt
& \reshl{86.11}{0.24} & \reshl{85.89}{0.43} & \reshl{85.20}{0.93}
& \reshl{76.96}{0.34} & \reshl{76.82}{0.04} & \reshl{76.71}{0.19}
& \reshl{84.39}{0.42} & \reshl{84.10}{0.19} & \reshl{83.91}{0.20}	
\\

FedSage
& \reshl{86.86}{0.15} & \reshl{86.59}{0.23} & \reshl{86.32}{0.37}	
& \reshl{77.91}{0.59} & \reshl{77.82}{0.13} & \reshl{77.30}{0.71}
& \reshl{87.75}{0.23} & \reshl{87.51}{0.20} & \reshl{87.49}{0.09}	
\\

\cline{2-10} 
\hline
\rowcolor[HTML]{FFF0C1}
\oursabbrv{}{}
& \tworowreshl{\ourreshl{\textbf{88.34}}{0.34}}{1.71} & \tworowreshl{\ourreshl{\textbf{88.56}}{0.43}}{2.35} & \tworowreshl{\ourreshl{\textbf{88.01}}{0.26}}{2.00}	
& \tworowreshl{\ourreshl{\textbf{80.43}}{0.23}}{4.06} & \tworowreshl{\ourreshl{\textbf{80.21}}{0.11}}{3.64} & \tworowreshl{\ourreshl{\textbf{80.01}}{0.09}}{4.09}
& \tworowreshl{\ourreshl{\textbf{88.25}}{0.60}}{2.96} & \tworowreshl{\ourreshl{\textbf{87.75}}{0.41}}{3.48} & \tworowreshl{\ourreshl{\textbf{87.60}}{0.53}}{2.73}	
\end{tabular}}}
\vspace{-5pt}
\captionsetup{font=small}
\caption{\small{
\textbf{Comparison with the state-of-the-art methods} on Cora, Citeseer and Pubmed datasets. The best result is bolded. {\selectfont\color{DarkBlue}{$\uparrow$}} means improved accuracy compared with FedAvg. {\selectfont\color{DarkGreen}{$\pm$}}  presents the standard deviation. Please see details in \cref{p:performance}.}}
\vspace{-5pt}
\label{table:performance}
\end{table*}

%settings

\noindent \textbf{Datasets}.
For node classification, our experiments are conducted on three benchmark datasets for the citation networks:
\begin{itemize}[leftmargin=*]
	\setlength{\itemsep}{0pt}
	\setlength{\parsep}{-2pt}
	\setlength{\parskip}{-0pt}
	\setlength{\leftmargin}{-10pt}
	\vspace{-4pt}
	
	\item \textbf{Cora} \cite{cora_IR20} dataset consists of 2708 scientific publications classified into one of seven classes. There are 5429 edges in the network of citations. 1433 distinct words make up the dictionary.
	
	\item \textbf{Citeseer} \cite{citeseer_98} dataset consists of 3312 scientific publications classified into one of six classes and 4732 edges. The dictionary contains 3703 unique words.

	\item \textbf{Pubmed} \cite{pubmed_08} dataset consists of 19717 scientific papers on diabetes that have been categorized into one of three categories in the PubMed database. The citation network has 44338 edges in it. A word vector from a dictionary with 500 unique terms that is TF/IDF weighted is used to describe each publication in the dataset.
 
	\vspace{-4pt}
\end{itemize}

\noindent \textbf{Network Structure}. Since the \gat{} \cite{GAT_arxiv17} is a powerful and widely used benchmark network in graph representing learning, we realize two layers \gat{} with parameter $\theta$, decoupling it into feature extractor $G(\cdot)$ and unified classifier $F(\cdot)$.
The hidden dimensions are 128 for all datasets, and classifier $F$ maps the embedding from 128 dimensions to 7,6,3 dimensions, which is the number of classification classes for Cora, Citeseer, and Pubmed respectively.

\noindent \textbf{Graph Augmentation Strategy}. 
Generating views is a key component of contrastive learning methods. In the graph domain, different views of a graph provide different contexts for each node. We follow augmentation mentioned in \cite{gcl-survey_arxiv21}, \cite{GRACE_ICML20} to construct a contrastive learning scheme. In \oursabbrv{}, we leverage two methods for new graph view generation, removing edges for topology and masking features for node attributes.
\begin{itemize}[leftmargin=*]
	\setlength{\itemsep}{0pt}
	\setlength{\parsep}{-2pt}
	\setlength{\parskip}{-0pt}
	\setlength{\leftmargin}{-10pt}
	\vspace{-4pt}
	
	\item \textbf{Removing edges (RE)}. It randomly removes a portion of edges in the original graph. 
	
	\item \textbf{Masking node features (MF)}. It randomly masks a fraction of dimensions with zeros in node features.
 
	\vspace{-4pt}
\end{itemize}

\begin{figure}[t]
	\begin{center}
		\includegraphics[width=\linewidth]{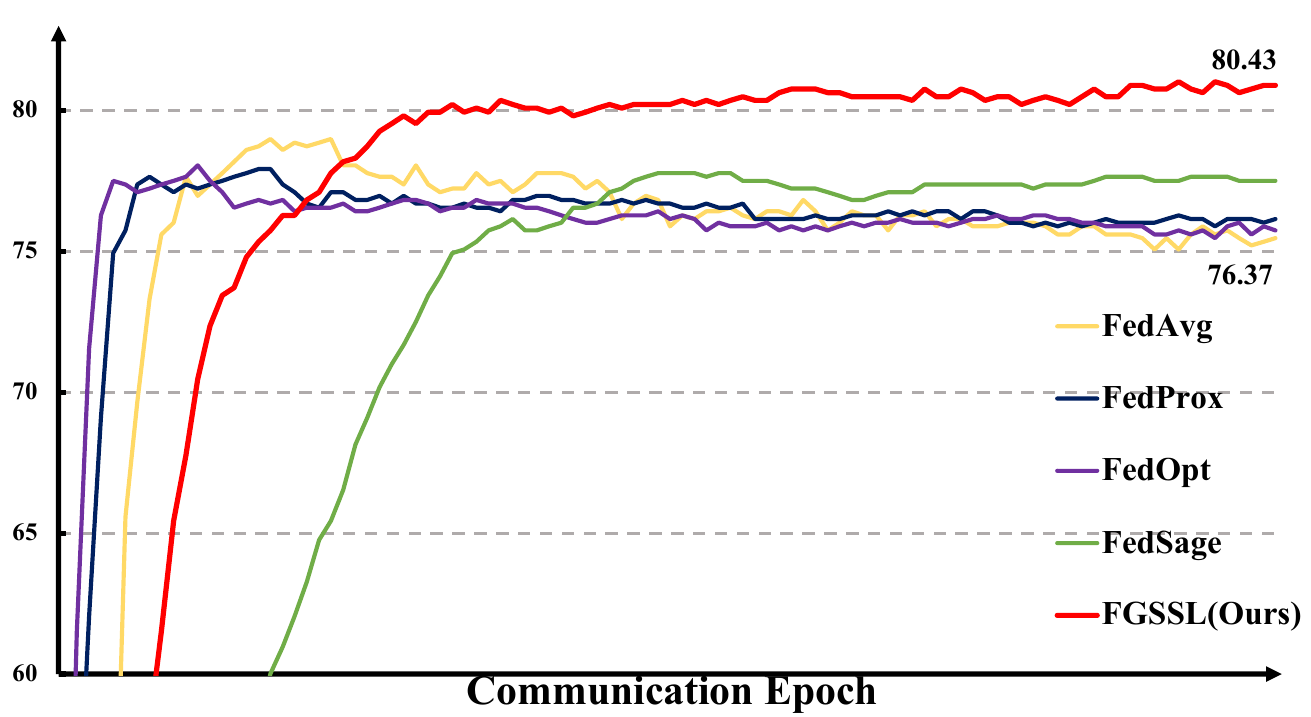}
		%\put(-235,5){\footnotesize (a) OCR}
		%\put(-113,5){\footnotesize (b) Ours}
	\end{center}
	\vspace{-10pt}
	\captionsetup{font=small}
	\caption{\small \textbf{Visualization of training curves} of the average test accuracy with Communication Epochs 200 with Citeseer dataset. Please see \cref{p:convergence} for details.
 }
	%\caption{\small Visualization of features learned by OCR~\cite{yuan2020object} and Ours on Cityscapes, under the supervision of $\mathcal{L}^{\text{CE}}$ (Eq.~\eqref{eq:CE}) and $\mathcal{L}^{\text{SEG}}$ (Eq.~\eqref{eq:com}), respectively. The points with different colors denote features from different classes. Our approach leads to a well-structured semantic space, showing better intra-class compactness and inter-class separability.}
	\vspace{-5pt}
	\label{fig:convergence}
\end{figure}

\noindent \textbf{Implement Details}. 
We utilize the community detection algorithm: Louvain, to simulate the subgraph systems. To stimulate the non-iid scene, this algorithm partitions the graph into multiple clusters and then assigns them to distributed clients. To conduct the experiments uniformly and fairly, we split the nodes into train/valid/test sets, where the ratio is 60\% : 20\% : 20\% . As for all networks, we use \textsl{SGD} \cite{SGD_AoMS51} as the selected optimizer with momentum $0.9$ and weight decay $5e-4$. The communication round is $200$ and the local training epoch is $4$ for all datasets. The metric used in our experiments is the node classification accuracy on the testing nodes and we report the averaged accuracy and the standard deviation over several random repetitions.

\noindent \textbf{Counterparts}.
(1) \textbf{Local}  each client train their model locally,
(2) \textbf{Global} the server leverage the complete graph for training.
For rigorous evaluation, we compare our \oursabbrv{} against popular federated strategies in \fgl{} setting.
(3) \textbf{FedAvg} (AISTATS'17 \cite{FedAvg_AISTATS17}),
(4) \textbf{FedProx} (MLSys'21 \cite{fedprox}),
(5) \textbf{FedOpt} (ICLR'21 \cite{FedOPT_ICLR21}), 
(6) \textbf{FedSage} (NeurIPS'21 \cite{fedsage+}).

\subsection{Experimental Results}
\noindent \textbf{Performance Comparison.}\label{p:performance}
The results of federated node classification for various methods under three non-IID settings are presented in \cref{table:performance}. These results indicate that \oursabbrv{} outperforms all other baselines and demonstrates a significant and consistent improvement compared to the conventional FedAvg algorithm in the \fgl{} setting. Additionally, personalized \fl{} algorithms such as FedProx and FedOpt demonstrate better performance than vanilla aggregation by utilizing a universal solution to the non-IID problem. Specialized methods in the \fgl{} field such as FedSage also perform better than common baselines, which is achieved through the simultaneous training of generative models for predicting missing links.

\noindent \textbf{Convergence Analysis.} \label{p:convergence}
\cref{fig:convergence} shows curves of the average test accuracy during the training process across five random runs conducted on the Citeseer datasets. It can be observed that \oursabbrv{} dominates the other methods in non-IID setting on the average test accuracy and achieves a stable convergence.

\subsection{Ablation Study}
\label{sec:ablation}
\begin{figure}[t]
\centering
\subfigure[Contrastive parameter $\tau$]
{	
\includegraphics[width=0.474\linewidth]{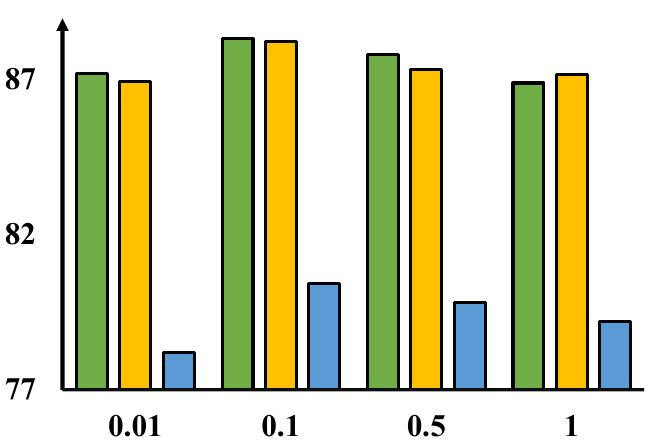}
}
\subfigure[Distillation parameter $\omega$]
{	
\includegraphics[width=0.474\linewidth]{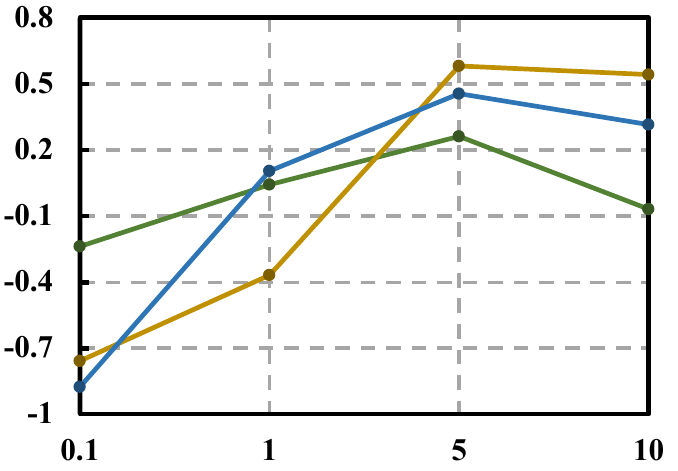}
}
\vspace{-10pt}
\captionsetup{font=small}
\caption{\small\textbf{Analysis on hyper-parameter in \oursabbrv{}}. Node classification results on three datasets under different $\tau$ and $\omega$ values with M = 5, in which green represents Cora, yellow represents Pubmed, and blue represents Citeseer.
Refer to \cref{p:hyper_parameter} for details.
}
\label{fig:hyper_parameter}
 \vspace{-5pt}
\end{figure}

% \begin{figure}[t]
% \centering
% \subfloat[Hyper-parameter study for $\tau$ with $\omega=5$.\label{fig:tau}]{\includegraphics[width=2.7in]{file/ablation_tau1.pdf}}\ \ \
% \subfloat[Hyper-parameter study for $\omega$ with $\tau=0.1$.\label{fig:omega}]{\includegraphics[width=2.7in]{file/abalation_tau2.pdf}}\ \ \
% \caption{Node classification accuracy on benchmark datasets under different $\tau$ and $\omega$ values with $M$=5} \label{fig:hyper-parameter}
% \end{figure}

\begin{table}[t]\small
\centering
{
\resizebox{\columnwidth}{!}{
		\setlength\tabcolsep{0.5pt}
		\renewcommand\arraystretch{1.2}
\begin{tabular}{c|c||cccIccc}
\hline \thickhline
\rowcolor{mygray}
&
& \multicolumn{3}{cI}{Cora} 
& \multicolumn{3}{c}{Citeseer}
\\
\cline{3-8} 
\rowcolor{mygray}
\multirow{-2}{*}{\firstmoduleabbrv{}}& \multirow{-2}{*}{\secondmoduleabbrv{}} 
& $M \! = \! 5$ & $M \! = \! 7$ & $M \! = \! 10$ 
& $M \! = \! 5$ & $M \! = \! 7$ & $M \! = \! 10$ 

\\

\hline\hline

\ding{55} & \ding{55}
& 86.63 & 86.21 & 86.01
& 76.37 & 76.57 & 75.92\\ 

\ding{55} & \ding{51}
& 86.86 & 86.32 & 86.51
& 77.91 & 77.53 & 76.42
\\

\ding{51}& \ding{55} 
& 88.01 & 88.23 & 87.84 
& 79.89 & 79.43 & 79.12\\ 

\rowcolor[HTML]{D7F6FF}
\ding{51} & \ding{51}
& \textbf{88.34} & \textbf{88.56} & \textbf{88.01} 
& \textbf{80.43} & \textbf{80.21} & \textbf{80.01}

\end{tabular}}}

\vspace{-5pt}
\captionsetup{font=small}
\caption{\small{
\textbf{Ablation study of key components} of our method in Cora and Citeseer datasets with clients 5/7/10. See \cref{p:key_components} for details.
}}
\label{table:key_component}
\vspace{-8pt}
\end{table}

\noindent \textbf{Effects of Key Components Mechanism}.
\label{p:key_components}
To better understand the impact of specific design components on the overall performance of \oursabbrv{}, we conducted an ablation study in which we varied these components.  For the variant without \firstmoduleabbrv{} and \secondmoduleabbrv{}, we utilize the vanilla \fgl{} setting with 2-layer \gat{}. As shown in \cref{table:key_component}, by exploiting both components, the best performance is achieved in all three graph datasets. It also suggests that \firstmoduleabbrv{} plays a more crucial role than \secondmoduleabbrv{}, which means the calibration in node semantics is stronger than the calibration in graph structure, and feature heterogeneity is more serious than graph heterogeneity in non-IID setting. Moreover, the contribution made by \secondmoduleabbrv{} is still not negligible and can benefit the learning process.

\noindent \textbf{Hyper-parameter study}.
\label{p:hyper_parameter}
 We compare the downstream task performance under different $\tau$ and $\omega$ values with five clients. Results are shown in \cref{table:performance}, where \cref{fig:hyper_parameter}(a) shows results when $\omega$ is fixed at 5, and \cref{fig:hyper_parameter}(b) shows results under $\tau=0.1$. It indicates that choosing $\tau$ can affect the strength of the contrastive method, where a smaller temperature benefits training more than higher ones, but extremely low temperatures (0.01) are harder to train due to numerical instability. Across different datasets, the optimal $\tau$ is constantly around 0.1. For choosing an appropriate $\tau$ in (\cref{eq:simdist_global} and \cref{eq:simdist_local}.), we find that the performance is not influenced much unless $\omega$ is set to extreme values like 0.1. 

\noindent \textbf{Discussion on Augmentation Strategies}.
\label{p:augmentation}
As demonstrated in \cref{table:augmentation}, different augmentation strategies were implemented within the augmentation module of proposed method. The experimental results indicate that utilizing two levels of augmentation improves performance. Specifically, on the one hand, using double-weak augmentation strategies did not result in a significant improvement when compared to baseline methods. On the other hand, double-strong augmentation strategies led to improved results as they allowed for exploration of rich semantic information through the supervised contrastive method. Additionally, the combination of strong and weak augmentation strategies at local and global levels, respectively, resulted in the highest overall performance, in accordance with our descriptions of them in \cref{sec:method}.

\begin{figure}[t]
\centering
\subfigure[\small FedAvg]
{	
\includegraphics[width=0.3\linewidth]{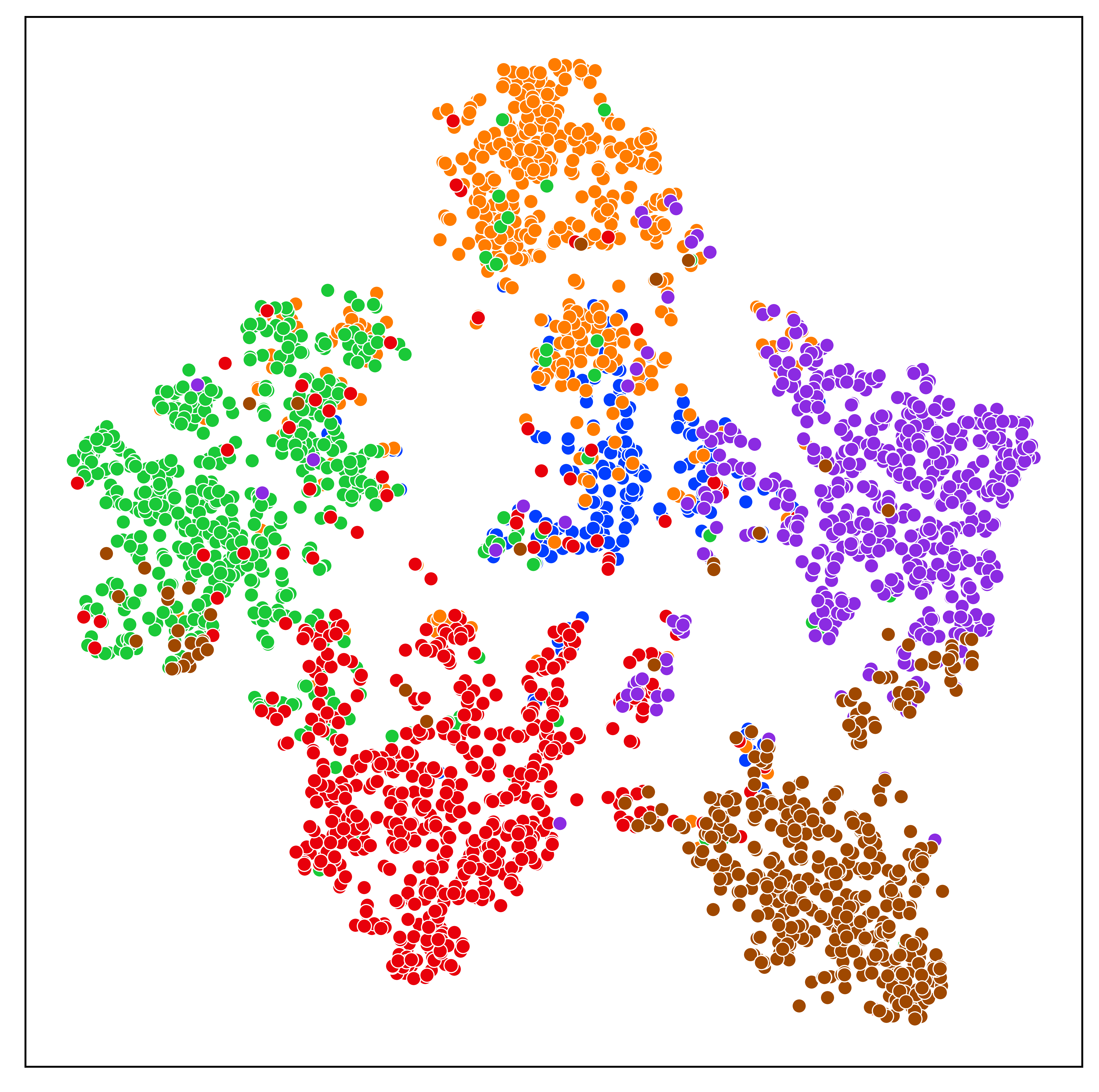}
}
\subfigure[\small FedSage]
{	
\includegraphics[width=0.3\linewidth]{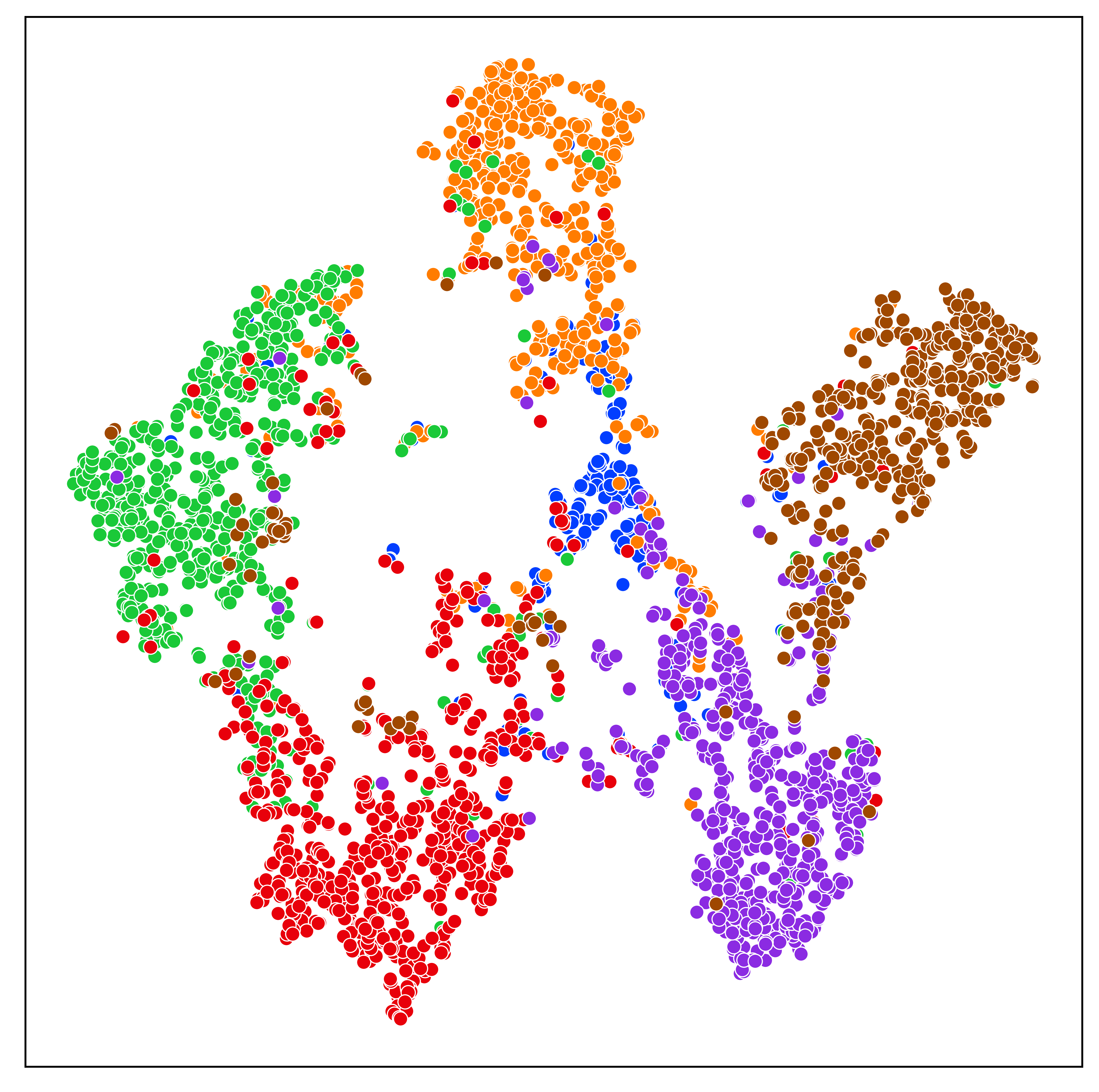}
}
\subfigure[\small FGSSL]
{	
\includegraphics[width=0.3\linewidth]{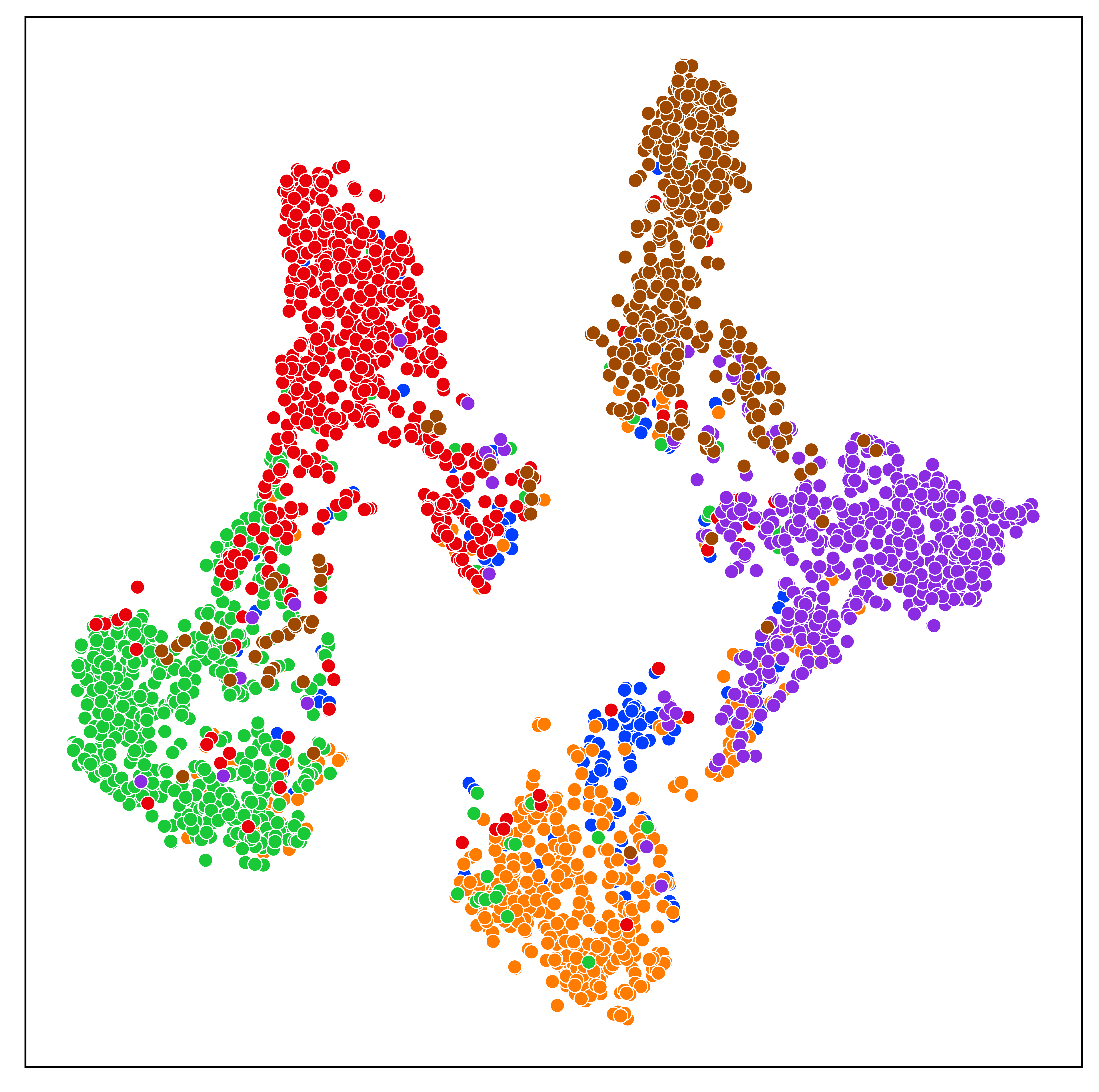}
}
\vspace{-5pt}
\captionsetup{font=small}
\caption{\small \textbf{Visualization of classification result.} The figure number corresponds to the method on the Citeseer dataset with $m=5$. Logits are colored based on class labels.
}
\label{fig:cka}
% \vspace{-5pt}
\end{figure}

\begin{table}[t]\small
\centering
{
\resizebox{\columnwidth}{!}{
		\setlength\tabcolsep{0.5pt}
		\renewcommand\arraystretch{1.2}
\begin{tabular}{cc||cccIccc}
\hline \thickhline
\rowcolor{mygray}
&
& \multicolumn{3}{cI}{Cora} 
& \multicolumn{3}{c}{Citeseer}
\\
\cline{3-8} 
\rowcolor{mygray}
\multirow{-2}{*}{Local}& \multirow{-2}{*}{Global} 
& $M \! = \! 5$ & $M \! = \! 7$ & $M \! = \! 10$ 
& $M \! = \! 5$ & $M \! = \! 7$ & $M \! = \! 10$ 

\\

\hline\hline

weak & weak
& 87.24 & 87.10 & 86.99
& 77.72 & 77.45 & 77.30
\\ 

weak & strong
& 86.86 & 86.68 & 86.48
& 77.22 & 77.09 & 76.33
\\

strong& strong 
& 87.91 & 87.93 & 87.52
& 79.59 & 79.12 & 78.81\\ 

\rowcolor[HTML]{D7F6FF}
strong & weak
& \textbf{88.01} & \textbf{88.23} & \textbf{87.84} 
& \textbf{79.89} & \textbf{79.43} & \textbf{79.12}

\end{tabular}}}

\vspace{-7pt}
\captionsetup{font=small}
\caption{\small{
\textbf{Analysis on augmentation strategies :} Effect of using weak or strong augmentations for two datasets trained on the sole \firstmoduleabbrv{} component with 200 epochs.
See \cref{sec:ablation} for details.
}}
\label{table:augmentation}
\vspace{-10pt}
\end{table}

% \noindent \textbf{Varying Local Training Epochs}.

\section{Conclusion}
In this paper, we propose a novel federated graph learning framework, namely \oursabbrv{}, that mitigates the non-IID 
issues via appropriately calibrating the heterogeneity both on the node-level semantic and graph-level structure. We develop two key components to solve the problems respectively. On the one hand, we leverage the contrastive-based method to correct the drift node semantics from the global ones that have Identical semantic information and achieve a high level of semantic discrimination at node level. On the other hand, we consider transforming adjacency relationships into a similarity distribution and utilizing a global model to distill this information into the local model, which maintains the structural information and corrects the structure heterogeneity. Experimental results illustrate that \oursabbrv{} consistently outperforms the state-of-the-art methods in federated graph scenarios.

\section*{Acknowledgments}
This work is partially supported by 
National Natural Science Foundation of China under Grant (62176188, 62225113), 
the Key Research and Development Program of Hubei Province (2021BAA187),
Zhejiang lab (NO.2022NF0AB01),
CCF-Huawei Populus Grove Fund (CCF-HuaweiTC2022003),
the Special Fund of Hubei Luojia Laboratory (220100015) and 
the Science and Technology Major Project of Hubei Province (Next-Generation AI Technologies) under Grant (2019AEA170).

%% file: appendix.tex
% \paragraph{Roadmap of Appendix} The Appendix is organized as follows.
% We list the notations table in Section~\ref{app:notation}.
% We provide theoretical proof of convergence in Section~\ref{app:proof}. 
% The algorithm of FedBN is described in Section~\ref{app:algorithm}.
\section*{APPENDIX}

\input{notations.tex}

%% file: notations.tex
\section{Notation Table}
\label{app:notation}
		
\begin{table}[h]\small
\centering
\scriptsize{\resizebox{0.9\linewidth}{!}{
\begin{tabular}{cl}
\hline
Notations & Description \\ \hline
$h$ & node-level feature, $x \in \mathbb{R}^d$ \\ 
$z$ & output logit \\
$d$ & dimension of $ h$ \\
$c_i$ & label \\
$\mathcal{C}$ & classes collection \\
$v_i$ & the node with indicator $i$ in graph \\
$A_i$ & the neighborhoods of the node $v_i$ \\
$j$ & the neighbors indicator in $\mathcal{A}_i$ \\
$A$ & adjacency matrix \\
$l$ & the layer of the model \\
$G(\cdot)$ & the \gnn{} feature extractor \\
$F(\cdot)$ & the \gnn{} classifier head \\
$M$ & total number of clients \\
$m$ & the index of client \\
$D^m$ & private data of the  $m$ client\\
$R$ & communication round \\
$\theta^m$ & model parameter of client $m$ \\
$E$ & local training epochs \\
$\sigma(\cdot)$ & ReLU activation function, $\sigma(\cdot) = \max\{\cdot,0\}$ \\
$\tau$ & the parameter in contrastive method \\
$\omega$ & the temperature in knowledge distillation \\
$Aug(\cdot)$ & the augmentation module for graph\\
$Aug_s(\cdot)$ & the stronger augmentation module for graph\\
$Aug_w(\cdot)$ & the weaker augmentation module for graph\\
$\widetilde{\mathcal{G}}$ & augmented graph \\
$g$ & the global signal \\
$\lambda_C$ & the $L^{FNSC}_i$ loss weight \\
$\lambda_D$ & the $L^{FGSD}_i$ loss weight \\
$\boldsymbol{P}$ & the collection of the positive indicators  \\
$\boldsymbol{K}$ & the collection of the negative indicators  \\
$p$ & the indicator of positive sample \\
$k$ & the indicator of negative sample \\
$S$ & the similarity of the selected node vector \\
$\eta$ & learning rate  \\
\hline
\end{tabular}%
}}

\caption{Notations occurred in the paper.}
\label{tab:notation}
\end{table}

%% file: ijcai23.bbl
\begin{thebibliography}{}

\bibitem[\protect\citeauthoryear{Baek \bgroup \em et al.\egroup }{2022}]{community_arxiv22}
Jinheon Baek, Wonyong Jeong, Jiongdao Jin, Jaehong Yoon, and Sung~Ju Hwang.
\newblock Personalized subgraph federated learning.
\newblock {\em arXiv preprint arXiv:2206.10206}, 2022.

\bibitem[\protect\citeauthoryear{Chen \bgroup \em et al.\egroup }{2020}]{SimCLR_ICML20}
Ting Chen, Simon Kornblith, Mohammad Norouzi, and Geoffrey Hinton.
\newblock A simple framework for contrastive learning of visual representations.
\newblock In {\em ICML}, pages 1597--1607, 2020.

\bibitem[\protect\citeauthoryear{Fang \bgroup \em et al.\egroup }{2021}]{SEED_ICLR21}
Zhiyuan Fang, Jianfeng Wang, Lijuan Wang, Lei Zhang, Yezhou Yang, and Zicheng Liu.
\newblock Seed: Self-supervised distillation for visual representation.
\newblock In {\em ICLR}, 2021.

\bibitem[\protect\citeauthoryear{Fu \bgroup \em et al.\egroup }{2022}]{fgl-survey_arxiv22}
Xingbo Fu, Binchi Zhang, Yushun Dong, Chen Chen, and Jundong Li.
\newblock Federated graph machine learning: A survey of concepts, techniques, and applications.
\newblock {\em arXiv preprint arXiv:2207.11812}, 2022.

\bibitem[\protect\citeauthoryear{Giles \bgroup \em et al.\egroup }{1998}]{citeseer_98}
C~Lee Giles, Kurt~D Bollacker, and Steve Lawrence.
\newblock Citeseer: An automatic citation indexing system.
\newblock In {\em Proceedings of the third ACM conference on Digital libraries}, pages 89--98, 1998.

\bibitem[\protect\citeauthoryear{Grover and Leskovec}{2016}]{node2vec_kdd16}
Aditya Grover and Jure Leskovec.
\newblock node2vec: Scalable feature learning for networks.
\newblock In {\em ACM SIGKDD}, pages 855--864, 2016.

\bibitem[\protect\citeauthoryear{Hassani and Khasahmadi}{2020}]{MVGRL_ICML20}
Kaveh Hassani and Amir~Hosein Khasahmadi.
\newblock Contrastive multi-view representation learning on graphs.
\newblock In {\em ICML}, pages 4116--4126, 2020.

\bibitem[\protect\citeauthoryear{He \bgroup \em et al.\egroup }{2016}]{ResNet_CVPR16}
Kaiming He, Xiangyu Zhang, Shaoqing Ren, and Jian Sun.
\newblock Deep residual learning for image recognition.
\newblock In {\em CVPR}, pages 770--778, 2016.

\bibitem[\protect\citeauthoryear{He \bgroup \em et al.\egroup }{2020}]{moco_CVPR20}
Kaiming He, Haoqi Fan, Yuxin Wu, Saining Xie, and Ross Girshick.
\newblock Momentum contrast for unsupervised visual representation learning.
\newblock In {\em CVPR}, pages 9729--9738, 2020.

\bibitem[\protect\citeauthoryear{Hinton \bgroup \em et al.\egroup }{2015}]{KD_arXiv15}
Geoffrey Hinton, Oriol Vinyals, and Jeff Dean.
\newblock Distilling the knowledge in a neural network.
\newblock {\em arXiv preprint arXiv:1503.02531}, 2015.

\bibitem[\protect\citeauthoryear{Hu \bgroup \em et al.\egroup }{2022}]{fedgcn_MDPI22}
Kai Hu, Jiasheng Wu, Yaogen Li, Meixia Lu, Liguo Weng, and Min Xia.
\newblock Fedgcn: Federated learning-based graph convolutional networks for non-euclidean spatial data.
\newblock {\em Mathematics}, 10(6):1000, 2022.

\bibitem[\protect\citeauthoryear{Huang \bgroup \em et al.\egroup }{2022a}]{FCCL_CVPR22}
Wenke Huang, Mang Ye, and Bo~Du.
\newblock Learn from others and be yourself in heterogeneous federated learning.
\newblock In {\em CVPR}, 2022.

\bibitem[\protect\citeauthoryear{Huang \bgroup \em et al.\egroup }{2022b}]{FSMAFL_ACMMM22}
Wenke Huang, Mang Ye, Bo~Du, and Xiang Gao.
\newblock Few-shot model agnostic federated learning.
\newblock In {\em ACM MM}, pages 7309--7316, 2022.

\bibitem[\protect\citeauthoryear{Huang \bgroup \em et al.\egroup }{2023a}]{FPL_CVPR23}
Wenke Huang, Mang Ye, Zekun Shi, He~Li, and Bo~Du.
\newblock Rethinking federated learning with domain shift: A prototype view.
\newblock In {\em CVPR}, 2023.

\bibitem[\protect\citeauthoryear{Huang \bgroup \em et al.\egroup }{2023b}]{FLSurveyandBenchmarkforGenRobFair_arXiv23}
Wenke Huang, Mang Ye, Zekun Shi, Guancheng Wan, He~Li, Bo~Du, and Qiang Yang.
\newblock A federated learning for generalization, robustness, fairness: A survey and benchmark.
\newblock {\em arXiv}, 2023.

\bibitem[\protect\citeauthoryear{Kairouz \bgroup \em et al.\egroup }{2019}]{Advances_arXiv19}
Peter Kairouz, H~Brendan McMahan, Brendan Avent, Aur{\'e}lien Bellet, Mehdi Bennis, Arjun~Nitin Bhagoji, Kallista Bonawitz, Zachary Charles, Graham Cormode, Rachel Cummings, et~al.
\newblock Advances and open problems in federated learning.
\newblock {\em arXiv preprint arXiv:1912.04977}, 2019.

\bibitem[\protect\citeauthoryear{Khosla \bgroup \em et al.\egroup }{2020}]{SupCon_NeurIPS20}
Prannay Khosla, Piotr Teterwak, Chen Wang, Aaron Sarna, Yonglong Tian, Phillip Isola, Aaron Maschinot, Ce~Liu, and Dilip Krishnan.
\newblock Supervised contrastive learning.
\newblock volume~33, pages 18661--18673, 2020.

\bibitem[\protect\citeauthoryear{Kipf and Welling}{2016}]{GAE_arxiv16}
Thomas~N Kipf and Max Welling.
\newblock Variational graph auto-encoders.
\newblock {\em arXiv preprint arXiv:1611.07308}, 2016.

\bibitem[\protect\citeauthoryear{Kipf and Welling}{2017}]{GCN_ICLR17}
Thomas~N Kipf and Max Welling.
\newblock Semi-supervised classification with graph convolutional networks.
\newblock In {\em ICLR}, 2017.

\bibitem[\protect\citeauthoryear{Kornblith \bgroup \em et al.\egroup }{2019}]{CKA_ICML19}
Simon Kornblith, Mohammad Norouzi, Honglak Lee, and Geoffrey Hinton.
\newblock Similarity of neural network representations revisited.
\newblock In {\em ICML}, pages 3519--3529, 2019.

\bibitem[\protect\citeauthoryear{Li \bgroup \em et al.\egroup }{2020}]{fedprox}
Tian Li, Anit~Kumar Sahu, Manzil Zaheer, Maziar Sanjabi, Ameet Talwalkar, and Virginia Smith.
\newblock Federated optimization in heterogeneous networks.
\newblock {\em Proceedings of Machine Learning and Systems}, 2:429--450, 2020.

\bibitem[\protect\citeauthoryear{Li \bgroup \em et al.\egroup }{2021}]{MOON_CVPR21}
Qinbin Li, Bingsheng He, and Dawn Song.
\newblock Model-contrastive federated learning.
\newblock In {\em CVPR}, pages 10713--10722, 2021.

\bibitem[\protect\citeauthoryear{Liu \bgroup \em et al.\egroup }{2022a}]{gcl-survey_IEEE22}
Yixin Liu, Ming Jin, Shirui Pan, Chuan Zhou, Yu~Zheng, Feng Xia, and Philip Yu.
\newblock Graph self-supervised learning: A survey.
\newblock {\em IEEE TKDE}, 2022.

\bibitem[\protect\citeauthoryear{Liu \bgroup \em et al.\egroup }{2022b}]{SUBLIME_WWW22}
Yixin Liu, Yu~Zheng, Daokun Zhang, Hongxu Chen, Hao Peng, and Shirui Pan.
\newblock Towards unsupervised deep graph structure learning.
\newblock In {\em Proceedings of the ACM Web Conference 2022}, pages 1392--1403, 2022.

\bibitem[\protect\citeauthoryear{Liu \bgroup \em et al.\egroup }{2023a}]{GOOD-WSDM23}
Yixin Liu, Kaize Ding, Huan Liu, and Shirui Pan.
\newblock Good-d: On unsupervised graph out-of-distribution detection.
\newblock In {\em WSDM}, 2023.

\bibitem[\protect\citeauthoryear{Liu \bgroup \em et al.\egroup }{2023b}]{GREET_AAAI23}
Yixin Liu, Yizhen Zheng, Daokun Zhang, Vincent Lee, and Shirui Pan.
\newblock Beyond smoothing: Unsupervised graph representation learning with edge heterophily discriminating.
\newblock In {\em AAAI}, 2023.

\bibitem[\protect\citeauthoryear{Liu \bgroup \em et al.\egroup }{2024}]{liu2024review}
Zewen Liu, Guancheng Wan, B~Aditya Prakash, Max~SY Lau, and Wei Jin.
\newblock A review of graph neural networks in epidemic modeling.
\newblock {\em arXiv preprint arXiv:2403.19852}, 2024.

\bibitem[\protect\citeauthoryear{McCallum \bgroup \em et al.\egroup }{2000}]{cora_IR20}
Andrew~Kachites McCallum, Kamal Nigam, Jason Rennie, and Kristie Seymore.
\newblock Automating the construction of internet portals with machine learning.
\newblock {\em Information Retrieval}, 3(2):127--163, 2000.

\bibitem[\protect\citeauthoryear{McMahan \bgroup \em et al.\egroup }{2017}]{FedAvg_AISTATS17}
Brendan McMahan, Eider Moore, Daniel Ramage, Seth Hampson, and Blaise~Aguera y~Arcas.
\newblock Communication-efficient learning of deep networks from decentralized data.
\newblock In {\em AISTATS}, pages 1273--1282, 2017.

\bibitem[\protect\citeauthoryear{Mullapudi \bgroup \em et al.\egroup }{2019}]{modeldiss_ECCV19}
Ravi~Teja Mullapudi, Steven Chen, Keyi Zhang, Deva Ramanan, and Kayvon Fatahalian.
\newblock Online model distillation for efficient video inference.
\newblock In {\em ECCV}, pages 3573--3582, 2019.

\bibitem[\protect\citeauthoryear{Panagopoulos \bgroup \em et al.\egroup }{2021}]{pandemic_AAAI21}
George Panagopoulos, Giannis Nikolentzos, and Michalis Vazirgiannis.
\newblock Transfer graph neural networks for pandemic forecasting.
\newblock In {\em AAAI}, pages 4838--4845, 2021.

\bibitem[\protect\citeauthoryear{Park \bgroup \em et al.\egroup }{2019}]{RKD_CVPR19}
Wonpyo Park, Dongju Kim, Yan Lu, and Minsu Cho.
\newblock Relational knowledge distillation.
\newblock In {\em CVPR}, pages 3967--3976, 2019.

\bibitem[\protect\citeauthoryear{Perozzi \bgroup \em et al.\egroup }{2014}]{deepwalk_kdd14}
Bryan Perozzi, Rami Al-Rfou, and Steven Skiena.
\newblock Deepwalk: Online learning of social representations.
\newblock In {\em ACM SIGKDD}, pages 701--710, 2014.

\bibitem[\protect\citeauthoryear{Reddi \bgroup \em et al.\egroup }{2021}]{FedOPT_ICLR21}
Sashank Reddi, Zachary Charles, Manzil Zaheer, Zachary Garrett, Keith Rush, Jakub Kone{\v{c}}n{\`y}, Sanjiv Kumar, and H~Brendan McMahan.
\newblock Adaptive federated optimization.
\newblock In {\em ICLR}, 2021.

\bibitem[\protect\citeauthoryear{Robbins and Monro}{1951}]{SGD_AoMS51}
Herbert Robbins and Sutton Monro.
\newblock A stochastic approximation method.
\newblock {\em AoMS}, pages 400--407, 1951.

\bibitem[\protect\citeauthoryear{Romero \bgroup \em et al.\egroup }{2015}]{Fitnets_ICLR15}
Adriana Romero, Nicolas Ballas, Samira~Ebrahimi Kahou, Antoine Chassang, Carlo Gatta, and Yoshua Bengio.
\newblock Fitnets: Hints for thin deep nets.
\newblock In {\em ICLR}, 2015.

\bibitem[\protect\citeauthoryear{Sen \bgroup \em et al.\egroup }{2008}]{pubmed_08}
Prithviraj Sen, Galileo Namata, Mustafa Bilgic, Lise Getoor, Brian Galligher, and Tina Eliassi-Rad.
\newblock Collective classification in network data.
\newblock {\em AI magazine}, 29(3):93--93, 2008.

\bibitem[\protect\citeauthoryear{Tan \bgroup \em et al.\egroup }{2023}]{FedStar_AAAI23}
Yue Tan, Yixin Liu, Guodong Long, Jing Jiang, Qinghua Lu, and Chengqi Zhang.
\newblock Federated learning on non-iid graphs via structural knowledge sharing.
\newblock In {\em AAAI}, 2023.

\bibitem[\protect\citeauthoryear{Tejankar \bgroup \em et al.\egroup }{2021}]{ISD_ICCV21}
Ajinkya Tejankar, Soroush~Abbasi Koohpayegani, Vipin Pillai, Paolo Favaro, and Hamed Pirsiavash.
\newblock Isd: Self-supervised learning by iterative similarity distillation.
\newblock In {\em ICCV}, 2021.

\bibitem[\protect\citeauthoryear{Vaswani \bgroup \em et al.\egroup }{2017}]{transformer_NeurIPS17}
Ashish Vaswani, Noam Shazeer, Niki Parmar, Jakob Uszkoreit, Llion Jones, Aidan~N Gomez, {\L}ukasz Kaiser, and Illia Polosukhin.
\newblock Attention is all you need.
\newblock {\em NeurIPS}, 30, 2017.

\bibitem[\protect\citeauthoryear{Veli{\v{c}}kovi{\'c} \bgroup \em et al.\egroup }{2017}]{GAT_arxiv17}
Petar Veli{\v{c}}kovi{\'c}, Guillem Cucurull, Arantxa Casanova, Adriana Romero, Pietro Lio, and Yoshua Bengio.
\newblock Graph attention networks.
\newblock {\em arXiv preprint arXiv:1710.10903}, 2017.

\bibitem[\protect\citeauthoryear{Wan \bgroup \em et al.\egroup }{2024a}]{FGGP_AAAI24}
Guancheng Wan, Wenke Huang, and Mang Ye.
\newblock Federated graph learning under domain shift with generalizable prototypes.
\newblock In {\em Proceedings of the AAAI Conference on Artificial Intelligence}, volume~38, pages 15429--15437, 2024.

\bibitem[\protect\citeauthoryear{Wan \bgroup \em et al.\egroup }{2024b}]{S3GCL_ICML24}
Guancheng Wan, Yijun Tian, Wenke Huang, Nitesh~V Chawla, and Mang Ye.
\newblock S3gcl: Spectral, swift, spatial graph contrastive learning.
\newblock In {\em ICML}, 2024.

\bibitem[\protect\citeauthoryear{Wang and Yoon}{2021}]{KD_PAMI_21}
Lin Wang and Kuk-Jin Yoon.
\newblock Knowledge distillation and student-teacher learning for visual intelligence: A review and new outlooks.
\newblock {\em IEEE TPAMI}, 2021.

\bibitem[\protect\citeauthoryear{Xie \bgroup \em et al.\egroup }{2021}]{fedGCL_NeurIPS21}
Han Xie, Jing Ma, Li~Xiong, and Carl Yang.
\newblock Federated graph classification over non-iid graphs.
\newblock {\em NeurIPS}, 34:18839--18852, 2021.

\bibitem[\protect\citeauthoryear{Ye \bgroup \em et al.\egroup }{2019}]{mangye_cl_CVPR19}
Mang Ye, Xu~Zhang, Pong~C Yuen, and Shih-Fu Chang.
\newblock Unsupervised embedding learning via invariant and spreading instance feature.
\newblock In {\em CVPR}, pages 6210--6219, 2019.

\bibitem[\protect\citeauthoryear{Ye \bgroup \em et al.\egroup }{2022}]{mangye_agu_PAMI20}
Mang Ye, Jianbing Shen, Xu~Zhang, Pong~C. Yuen, and Shih-Fu Chang.
\newblock Augmentation invariant and instance spreading feature for softmax embedding.
\newblock {\em IEEE TPAMI}, 44(2):924--939, 2022.

\bibitem[\protect\citeauthoryear{Zhang \bgroup \em et al.\egroup }{2021}]{fedsage+}
Ke~Zhang, Carl Yang, Xiaoxiao Li, Lichao Sun, and Siu~Ming Yiu.
\newblock Subgraph federated learning with missing neighbor generation.
\newblock {\em NeurIPS}, 34:6671--6682, 2021.

\bibitem[\protect\citeauthoryear{Zhu \bgroup \em et al.\egroup }{2020}]{GRACE_ICML20}
Yanqiao Zhu, Yichen Xu, Feng Yu, Qiang Liu, Shu Wu, and Liang Wang.
\newblock Deep graph contrastive representation learning.
\newblock In {\em ICML}, 2020.

\bibitem[\protect\citeauthoryear{Zhu \bgroup \em et al.\egroup }{2021a}]{filt+_arxiv21}
Wei Zhu, Andrew White, and Jiebo Luo.
\newblock Federated learning of molecular properties in a heterogeneous setting.
\newblock {\em arXiv preprint arXiv:2109.07258}, 2021.

\bibitem[\protect\citeauthoryear{Zhu \bgroup \em et al.\egroup }{2021b}]{gcl-survey_arxiv21}
Yanqiao Zhu, Yichen Xu, Qiang Liu, and Shu Wu.
\newblock An empirical study of graph contrastive learning.
\newblock {\em arXiv preprint arXiv:2109.01116}, 2021.

\end{thebibliography}
